\documentclass{article}

\usepackage{microtype}
\usepackage{graphicx}
\usepackage{booktabs} 

\usepackage{hyperref}

\usepackage{subfig}

\usepackage[accepted]{icml2024}

\usepackage{amsmath}
\usepackage{amssymb}
\usepackage{mathtools}
\usepackage{amsthm}

\usepackage[capitalize,noabbrev]{cleveref}

\theoremstyle{plain}

\theoremstyle{definition}

\theoremstyle{remark}

\makeatletter
\providecommand\phantomcaption{\caption@refstepcounter\@captype}
\makeatother

\usepackage[textsize=tiny]{todonotes}
\usepackage[subtle]{savetrees}

\icmltitlerunning{Vid3D: Synthesis of Dynamic 3D Scenes using 2D Video Diffusion}

\begin{document}

\twocolumn[
\icmltitle{Vid3D: Synthesis of Dynamic 3D Scenes using 2D Video Diffusion}



\icmlsetsymbol{equal}{*}

\begin{icmlauthorlist}
\icmlauthor{Rishab Parthasarathy}{equal,yyy}
\icmlauthor{Zachary Ankner}{equal,yyy,zzz}
\icmlauthor{Aaron Gokaslan}{zzz,xxx}

\end{icmlauthorlist}

\icmlaffiliation{yyy}{Department of EECS, MIT, Cambridge, MA, USA}
\icmlaffiliation{zzz}{Databricks Mosaic AI, San Francisco, CA, USA}
\icmlaffiliation{xxx}{Department of Computer Science, Cornell, Ithaca, NY, USA}

\icmlcorrespondingauthor{Rishab Parthasarathy}{rpartha@mit.edu}
\icmlcorrespondingauthor{Zachary Ankner}{ankner@mit.edu}

\icmlkeywords{Machine Learning, ICML, Computer Vision, 3D Video, Video Generation}

\vskip 0.3in
]



\printAffiliationsAndNotice{\icmlEqualContribution} 


\begin{abstract}
A recent frontier in computer vision has been the task of 3D video generation, which consists of generating a time-varying 3D representation of a scene.
To generate dynamic 3D scenes, current methods explicitly model 3D temporal dynamics by jointly optimizing for consistency across both time and views of the scene.
In this paper, we instead investigate whether it is necessary to explicitly enforce multiview consistency over time, as current approaches do, or if it is sufficient for a model to generate 3D representations of each timestep independently.
We hence propose a model, Vid3D, that leverages 2D video diffusion to generate 3D videos by first generating a 2D "seed" of the video's temporal dynamics and then independently generating a 3D representation for each timestep in the seed video.
We evaluate Vid3D against two state-of-the-art 3D video generation methods and find that Vid3D is achieves comparable results despite not explicitly modeling 3D temporal dynamics. 
We further ablate how the quality of Vid3D depends on the number of views generated per frame.
While we observe some degradation with fewer views, performance degradation remains minor.
Our results thus suggest that 3D temporal knowledge may not be necessary to generate high-quality dynamic 3D scenes, potentially enabling simpler generative algorithms for this task.
\end{abstract}
\section{Introduction}
\label{sec:intro}

A key theme throughout the evolution of computer vision has been the modeling of new modalities that better capture the complexity of the real world~\citep{brooks2024sora, kerbl2023splatting, mildenhall2020nerf}.
A less explored and currently emerging frontier in computer vision is modelling dynamic 3D scenes instead of static 3D scenes.
In this work, we tackle the challenge of efficiently generating high-quality 3D videos.

Recent works have proposed methods for 3D video generation that fall into one of two paradigms.
In the first paradigm, classifier guidance from 2D video models and static 3D scene models is leveraged to optimize a dynamic 3D scene representation along multiple axes~\citep{zhao2023animate124, bahmani4dfy}.
In the second paradigm, a 3D representation is constructed for the first frame and then deformed over time to be consistent with a 2D rendering of the scene~\citep{rendreamgaussian}.
While these results achieve impressive results, they still have significant limitations.

Methods that rely on classifier guidance are extremely computationally intensive, requiring hours to generate a single 3D video. In contrast, methods that deform an initial 3D scene representation are more efficient, but they instead require rigid temporal structure.
Furthermore, both paradigms are very technically complex, introducing many hyperparameters that must be tuned.

\begin{figure*}[htbp]
    \centering
    \includegraphics[width=\textwidth]{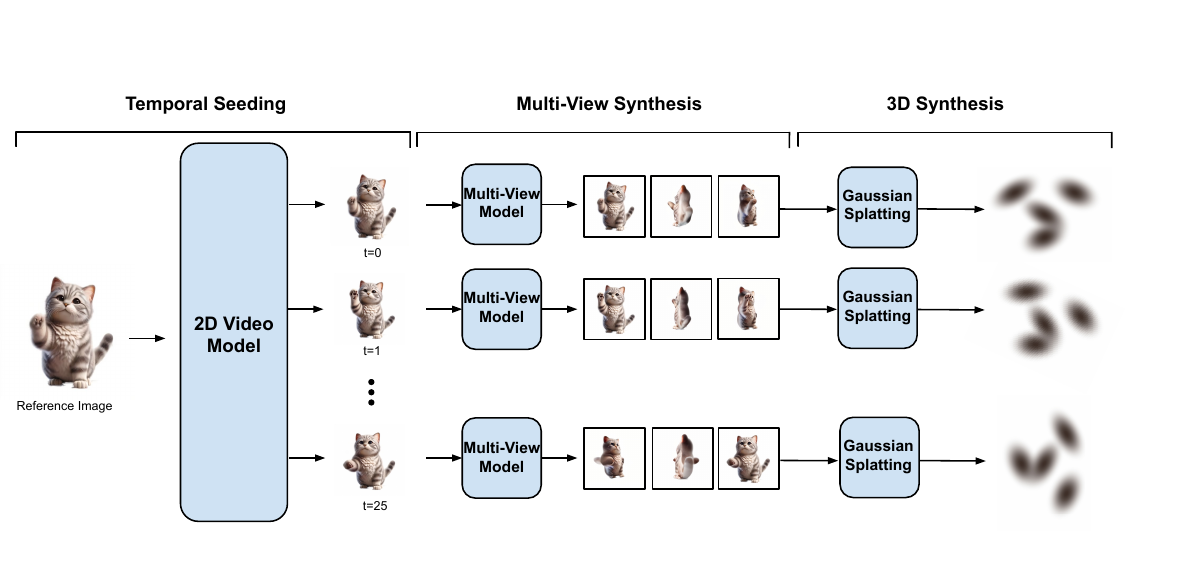}
    \caption{An overview of the Vid3D method. In stage 1, we generate a 2D video from a reference image to seed the dynamics of the scene. In stage 2, we generate multiple views for each timestep in the 2D video. In stage 3, we train a Gaussian splat on the collection of views from each timestep. Ultimately, each trained Gaussian splat represents a timestep in the 3D video.}
    \label{fig:method}
\end{figure*}

In this work, we thus investigate the common assumption that 3D temporal dynamics need to be explicitly modelled to perform 3D video generation.
Specifically, we consider whether a model can generate high-quality 3D videos without explicitly enforcing consistency between multiple views over time.
We propose Vid3D, a simple end-to-end pipeline for synthesizing dynamic 3D scenes from a single reference image.
We generate 3D videos in three steps: (1) generating a 2D video outline for the 3D video, (2) independently generating multiple views for each timestep in the seed video, and (3) using the multiple views to generate 3D representations of each frame.
We evaluate Vid3D against two state-of-the-art 3D video generation methods, finding that it achieves comparable quality even without explicitly modelling the 3D temporal dynamics of the scene.
We further ablate key parameters of Vid3D to gain a deeper understanding of the method.
Overall, we challenge the current assumption that methods require explicit temporal consistency to achieve SOTA results in the important task of 3D video generation~\citep{rendreamgaussian, zhao2023animate124}.





\section{Our Approach}
We hypothesize that it is possible to generate dynamic 3D scenes from a source image without explicitly modeling 3D temporal dynamics, meaning that it is not necessary  to enforce multi-view consistency across time.
To create a model capable of generating 3D videos without modeling 3D temporal dynamics, we factorize the task into generating the 2D temporal dynamics of the scene and then generating 3D representations of each timestep in the 2D scene.
Specifically, we propose a three-step process for generating 3D scenes from a source image: (1) seeding the scene dynamics with a 2D video, (2) synthesizing multiple views of the scene for each time-step (also known as a frame) in the seed video, and (3) generating a 3D representation of each time step based on the generated views.
We present a diagram of our Vid3D method in~\Cref{fig:method}.

To construct the dynamics of the scene, which we refer to as \emph{temporal seeding}, we query a 2D video model with the reference image.
This seeding provides us with a dynamic rendering of the object from a singular view.
We next transform the obtained 2D representation of the dynamic scene into a 3D representation.
As we do not enforce multi-view consistency across time, we generate the 3D representation of each frame in our 2D seed independently of all other frames.
To generate 3D representations of each 2D frame we follow the methodology proposed by \citet{chen2024v3d}.
For each seed frame, we generate multiple views of the target object by querying a 2D video model finetuned for multi-view generation on a curated subset of the Objaverse dataset~\citep{objaverse}.
For each frame, we then train a Gaussian Splat~\citep{kerbl20233d-gauss} on the collection of generated views.
The sequence of Gaussian Splats ultimately defines a 3D video.

\subsection{Model Choices}

We use the Stable Video Diffusion model~\citep{blattmann2023stable} to generate 2D temporal seeds.
The model generates seeds that consist of 25 frames.
The seed generations are conditioned on a "motion score" that controls the degree of motion in the video; our main results use a motion score of 120.
To generate multiple views for each timestep in the seed video, we use a version of Stable Video Diffusion fine-tuned for the multiview generation task on Objaverse~\citep{chen2024v3d, sv3d, objaverse}. We select this model over Zero 1-to-3 as it generates higher quality multi-views over a number of quantitative and qualitative metrics including human evaluation~\citep{zero123, chen2024v3d, sv3d}.
By default, this model generates 18 views that are uniformly sampled from a full orbit with zero elevation along the azimuth.
To obtain 3D representations from the collection of views, we train Gaussian splats~\citep{rendreamgaussian} consisting of 100K splats that are optimized for 4K steps.

\begin{figure*}[htbp]
    \centering
    \includegraphics[width=\textwidth]{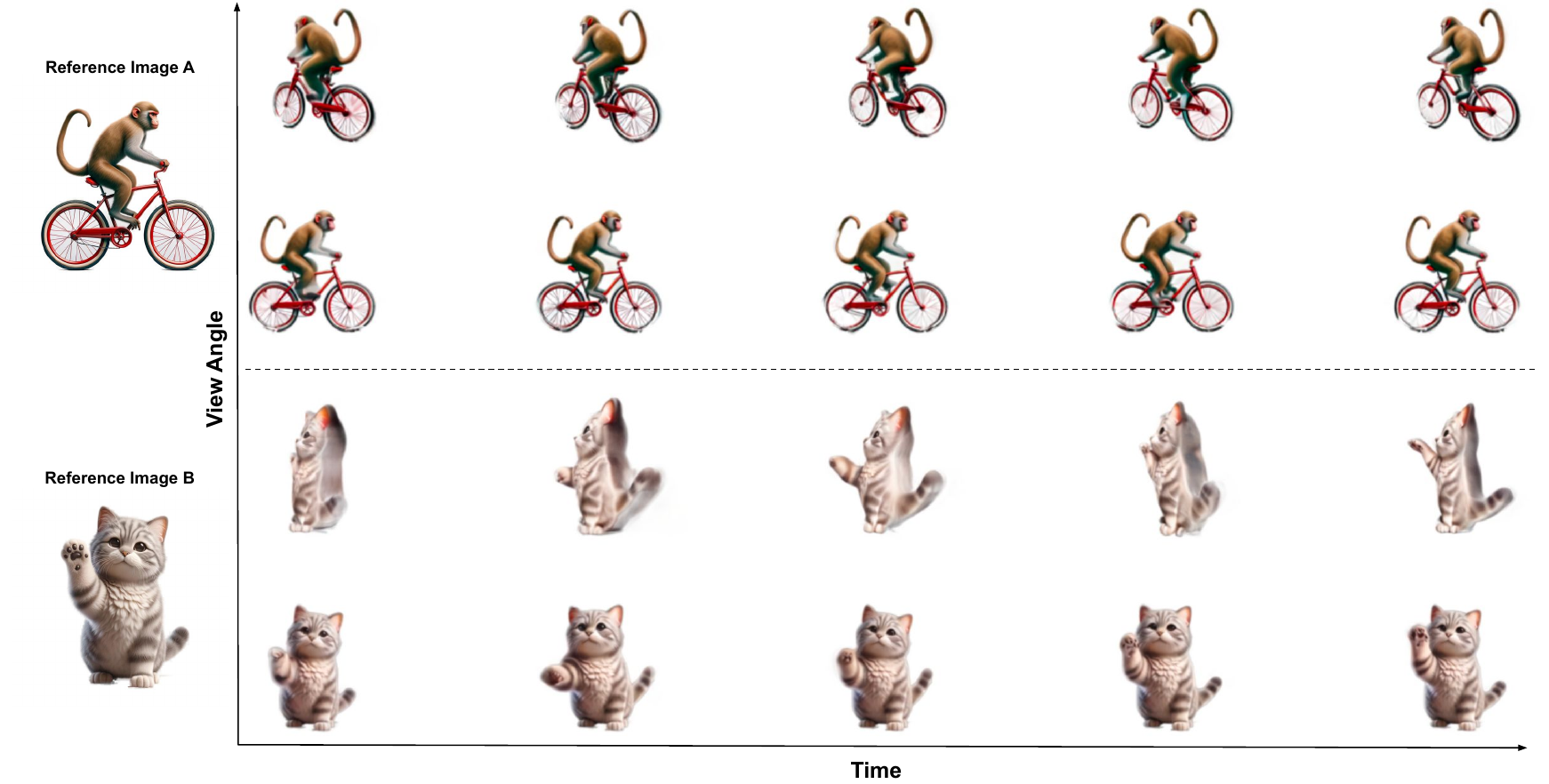}
    \caption{An example of multiple 2D renderings of 3D videos generated by our method. The 3D videos are rendered from two different camera views (y-axis) through time (x-axis). We observe consistency between different camera views for the same timestep as well as plausible dynamics across time for the same view.}
    \label{fig:main-result}
\end{figure*}

\subsection{Evaluation Setup}

To evaluate Vid3D we use the benchmark proposed by \citet{zhao2023animate124}, which is composed of 24 high-quality reference images.
This benchmark has been adopted by recent 3D video papers as the standard evaluation framework~\citep{zhao2023animate124, rendreamgaussian}.
Following their evaluation procedure, to measure the quality of each 3D video, we render 10 2D videos from ten different uniformly sampled camera angles. 
As a quantitative evaluation metric, we report the CLIP-I score, which is defined as the average cosine similarity between the CLIP-features~\citep{radford2021learning} of the reference image and each frame in each 2D video rendering. Specifically, we use the CLIP-ViT-Base with a patch size of 32 as the image encoder.
As CLIP features represent an embedding of an image in a semantically meaningful latent space, this metric captures the average semantic similarity (and hence model fidelity) across both time and camera angle.
We compare Vid3D to both Animate124 and DreamGaussian4D which represent the state-of-the-art for 3D video generation.

\section{Results and Discussion}

\subsection{Main Results}
We provide 2D video renderings from two 3D videos generated by Vid3D in~\Cref{fig:main-result}.
For each 3D video, we render 2D videos from two distinct camera views.
For both reference images, the renderings demonstrate subject consistency with the original frame as well as interesting and plausible dynamic details such as the cat raising and lowering its paws or the monkey pedaling the bike.
In addition to temporal consistency across time for the same view, there is spatial consistency for different views from the same timestep; the subject and the content agree across multiple views.
While the capacity of Vid3D for general 3D scene rendering is impressive, there are still weaknesses to the model.
As can be seen in the cat rendering, rendering quality degrades moving from views aligned with the reference image to alternative camera views.
We believe this decrease in quality is not due to multi-view information being dropped between timesteps, suggesting that the quality could be improved with higher quality multi-view synthesis models.

\begin{table}[h!]
\centering
\caption{CLIP-I score for Vid3D compared to Animate124 and DreamGaussian4D, showing that our model does not need 3D temporal dynamics to yield competitive results.}
\vskip 0.15in
\label{tab:main-results}
\begin{tabular}{lc}
\hline
\textbf{Model} & \textbf{CLIP-I} \\ \hline
Animate124 & 0.8544 \\
DreamGaussian4D & 0.9227 \\
Vid3D (Ours) & 0.8946 \\
\hline
\end{tabular}
\vskip -0.1in
\end{table}

We also quantitatively compare our proposed Vid3D method to both Animate124~\citep{zhao2023animate124} and DreamGaussian4D~\citep{rendreamgaussian} in~\Cref{tab:main-results}.
While Vid3D achieves a higher CLIP-I score than Animate124, it underperforms the CLIP-I score of DreamGaussian4D.
Although Vid3D does not achieve a new state-of-the-art score, that Vid3D's quantitative performance is competitive with state-of-the-art 3D video baselines implies that 3D temporal dynamics may not be fully necessary for generating dynamic 3D scenes.
Instead, it is possible to rely solely on 2D video model priors to ensure multi-view consistency through time, which with further tuning, could achieve even greater performance.

We present further qualitative results and an analysis of failure modes in ~\Cref{appendix:qualitative}.

\subsection{Varying Number of Views}

\begin{table}[h!]
\centering
\caption{CLIP-I score values for Vid3D for different numbers of views. This result shows that reducing the number of views from 18 to 9 does not significantly degrade performance, while further reduction does.}
\vskip 0.15in
\label{tab:ablate-views}
\begin{tabular}{lc}
\hline
\textbf{Number of views} & \textbf{CLIP-I} \\ \hline
3 & 0.8532 \\
9 & 0.8879 \\
18 (Baseline) & 0.8946 \\
\hline
\end{tabular}
\vskip -0.1in
\end{table}

We ablate how the number of views produced per timestep during the multi-view synthesis process affects the quality of the resulting 3D generations.
The number of views generated determines the information content available when training Gaussian splats and more views provides denser coverage of the scene.
While for capturing views of real objects, i.e., object views captured by sensors, more views corresponds to higher quality 3D representations, our work synthesizes views from a 2D video model and as such we investigate the effect of view number on quality.
We report the CLIP-I score for varying view numbers in~\Cref{tab:ablate-views}.
We find that the CLIP-I score strictly decreases as the number of views decreases.
Furthermore, the decrease in CLIP-I score grows larger at lower view numbers; there is only a drop of 0.0067 from 18 to 9 frames, but a drop of 0.0347 from 9 to 3 frames.
As compared to the baselines, even at 3 frames Vid3D is comparable to Animate124, only performing worse by 0.0012.

\begin{figure}[htbp]
    \centering
    \includegraphics[width=\linewidth]{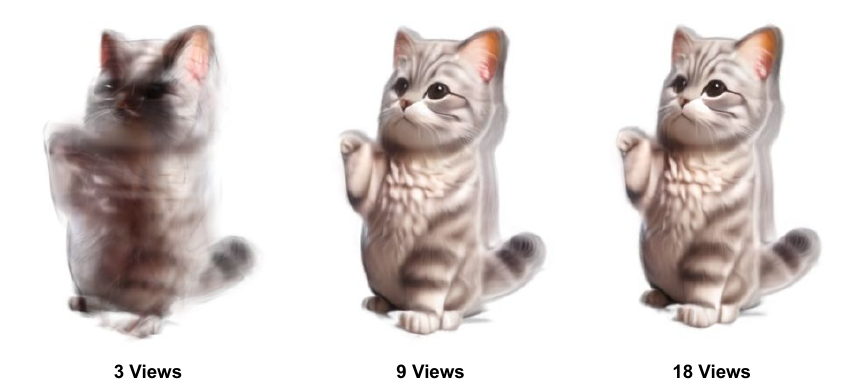}
    \caption{Rendering of singular frame from 3D videos generated using the same reference image but trained with a varying number of synthesized views. There are no perceivable differences between 18 and 9 views, but there is significant degradation and noise using 3 views.}
    \label{fig:ablate-views}
\end{figure}

We provide a qualitative example of single timestep renderings from Gaussian splats trained from varying number of views in~\Cref{fig:ablate-views}.
There is no perceivable degradation in quality when comparing 18 views to 9 views, but there is a significant loss of quality for the model trained with 3 views.
Specifically, at 3 views all high frequency detail is lost and the boundary of the object is poorly defined.

\section{Conclusion}
\label{sec:conclusion}

In this work, we systematically analyze an alternative to recent dynamic 3D scene generation algorithms and challenge the assumption that 3D video generation methods must explicitly model 3D temporal dynamic.
We propose Vid3D, a simple method for 3D video generation that generates 3D representations for each timestep in the video independently of all other timesteps.
We find that this simpler methodology still holds up to the state-of-the-art baselines of Animate124 and DreamGaussian4D.
Vid3D achieves a CLIP-I score of 0.8946 as compared to baseline scores of 0.8544 and 0.9227 for Animate124 and DreamGaussian4D respectively.
We ablate the effect that the number of views generated for each timestep has on 3D scene quality, finding that scene quality only significantly degrades for a very low number of views.
Overall, Vid3D is a simple alternative to current 3D video generation methods and it achieves results consistent with the state-of-the-art even though it does not explicitly model 3D temporal dynamics.

\section*{Impact Statement}

This paper presents work whose goal is to advance the field of 
Machine Learning. With easier generation of dynamic 3D scenes, embodied intelligence models can be trained more easily. This could result in societal benefits such as better trained robotics and prosthetics. However, we also elect to also point out the potential effects of this research on the creation of malicious videos like deepfakes. We encourage further work along preventing this malicious behavior for greater alignment of future models.


\bibliography{main}

\begin{thebibliography}{22}
\providecommand{\natexlab}[1]{#1}
\providecommand{\url}[1]{\texttt{#1}}
\expandafter\ifx\csname urlstyle\endcsname\relax
  \providecommand{\doi}[1]{doi: #1}\else
  \providecommand{\doi}{doi: \begingroup \urlstyle{rm}\Url}\fi

\bibitem[Armandpour et~al.(2023)Armandpour, Sadeghian, Zheng, Sadeghian, and Zhou]{janus}
Armandpour, M., Sadeghian, A., Zheng, H., Sadeghian, A., and Zhou, M.
\newblock Re-imagine the negative prompt algorithm: Transform 2d diffusion into 3d, alleviate janus problem and beyond.
\newblock \emph{CoRR}, abs/2304.04968, 2023.
\newblock \doi{10.48550/ARXIV.2304.04968}.
\newblock URL \url{https://doi.org/10.48550/arXiv.2304.04968}.

\bibitem[Bahmani et~al.(2023)Bahmani, Skorokhodov, Rong, Wetzstein, Guibas, Wonka, Tulyakov, Park, Tagliasacchi, and Lindell]{bahmani4dfy}
Bahmani, S., Skorokhodov, I., Rong, V., Wetzstein, G., Guibas, L.~J., Wonka, P., Tulyakov, S., Park, J.~J., Tagliasacchi, A., and Lindell, D.~B.
\newblock 4d-fy: Text-to-4d generation using hybrid score distillation sampling.
\newblock \emph{CoRR}, abs/2311.17984, 2023.
\newblock \doi{10.48550/ARXIV.2311.17984}.
\newblock URL \url{https://doi.org/10.48550/arXiv.2311.17984}.

\bibitem[Blattmann et~al.(2023)Blattmann, Dockhorn, Kulal, Mendelevitch, Kilian, Lorenz, Levi, English, Voleti, Letts, Jampani, and Rombach]{blattmann2023stable}
Blattmann, A., Dockhorn, T., Kulal, S., Mendelevitch, D., Kilian, M., Lorenz, D., Levi, Y., English, Z., Voleti, V., Letts, A., Jampani, V., and Rombach, R.
\newblock Stable video diffusion: Scaling latent video diffusion models to large datasets, 2023.

\bibitem[Brooks et~al.(2024)Brooks, Peebles, Holmes, DePue, Guo, Jing, Schnurr, Taylor, Luhman, Luhman, Ng, Wang, and Ramesh]{brooks2024sora}
Brooks, T., Peebles, B., Holmes, C., DePue, W., Guo, Y., Jing, L., Schnurr, D., Taylor, J., Luhman, T., Luhman, E., Ng, C., Wang, R., and Ramesh, A.
\newblock Video generation models as world simulators.
\newblock 2024.
\newblock URL \url{https://openai.com/research/video-generation/-models-as-world-simulators}.

\bibitem[Chang et~al.(2015)Chang, Funkhouser, Guibas, Hanrahan, Huang, Li, Savarese, Savva, Song, Su, Xiao, Yi, and Yu]{shapenet2015}
Chang, A.~X., Funkhouser, T., Guibas, L., Hanrahan, P., Huang, Q., Li, Z., Savarese, S., Savva, M., Song, S., Su, H., Xiao, J., Yi, L., and Yu, F.
\newblock {ShapeNet: An Information-Rich 3D Model Repository}.
\newblock Technical Report arXiv:1512.03012 [cs.GR], Stanford University --- Princeton University --- Toyota Technological Institute at Chicago, 2015.

\bibitem[Chatzitofis et~al.(2020)Chatzitofis, Saroglou, Boutis, Drakoulis, Zioulis, Subramanyam, Kevelham, Charbonnier, Cesar, Zarpalas, et~al.]{chatzitofis2020human4d}
Chatzitofis, A., Saroglou, L., Boutis, P., Drakoulis, P., Zioulis, N., Subramanyam, S., Kevelham, B., Charbonnier, C., Cesar, P., Zarpalas, D., et~al.
\newblock Human4d: A human-centric multimodal dataset for motions and immersive media.
\newblock \emph{IEEE Access}, 8:\penalty0 176241--176262, 2020.

\bibitem[Chen et~al.(2024)Chen, Wang, Wang, Wang, and Liu]{chen2024v3d}
Chen, Z., Wang, Y., Wang, F., Wang, Z., and Liu, H.
\newblock V3d: Video diffusion models are effective 3d generators.
\newblock \emph{arXiv preprint arXiv:2403.06738}, 2024.

\bibitem[Deitke et~al.(2023)Deitke, Schwenk, Salvador, Weihs, Michel, VanderBilt, Schmidt, Ehsani, Kembhavi, and Farhadi]{objaverse}
Deitke, M., Schwenk, D., Salvador, J., Weihs, L., Michel, O., VanderBilt, E., Schmidt, L., Ehsani, K., Kembhavi, A., and Farhadi, A.
\newblock Objaverse: A universe of annotated 3d objects.
\newblock In \emph{Proceedings of the IEEE/CVF Conference on Computer Vision and Pattern Recognition}, pp.\  13142--13153, 2023.

\bibitem[Du et~al.(2021)Du, Watkins, Darrell, Abbeel, and Pathak]{duautosim2real}
Du, Y., Watkins, O., Darrell, T., Abbeel, P., and Pathak, D.
\newblock Auto-tuned sim-to-real transfer.
\newblock In \emph{{IEEE} International Conference on Robotics and Automation, {ICRA} 2021, Xi'an, China, May 30 - June 5, 2021}, pp.\  1290--1296. {IEEE}, 2021.
\newblock \doi{10.1109/ICRA48506.2021.9562091}.
\newblock URL \url{https://doi.org/10.1109/ICRA48506.2021.9562091}.

\bibitem[Han et~al.(2024)Han, Kokkinos, and Torr]{han2024vfusion3d}
Han, J., Kokkinos, F., and Torr, P.
\newblock Vfusion3d: Learning scalable 3d generative models from video diffusion models, 2024.

\bibitem[Kerbl et~al.(2023{\natexlab{a}})Kerbl, Kopanas, Leimk{\"u}hler, and Drettakis]{kerbl20233d-gauss}
Kerbl, B., Kopanas, G., Leimk{\"u}hler, T., and Drettakis, G.
\newblock 3d gaussian splatting for real-time radiance field rendering.
\newblock \emph{ACM Transactions on Graphics}, 42\penalty0 (4):\penalty0 1--14, 2023{\natexlab{a}}.

\bibitem[Kerbl et~al.(2023{\natexlab{b}})Kerbl, Kopanas, Leimk{\"u}hler, and Drettakis]{kerbl2023splatting}
Kerbl, B., Kopanas, G., Leimk{\"u}hler, T., and Drettakis, G.
\newblock 3d gaussian splatting for real-time radiance field rendering.
\newblock \emph{ACM Transactions on Graphics}, 42\penalty0 (4), July 2023{\natexlab{b}}.
\newblock URL \url{https://repo-sam.inria.fr/fungraph/3d-gaussian-splatting/}.

\bibitem[Liu et~al.(2023)Liu, Wu, Hoorick, Tokmakov, Zakharov, and Vondrick]{zero123}
Liu, R., Wu, R., Hoorick, B.~V., Tokmakov, P., Zakharov, S., and Vondrick, C.
\newblock Zero-1-to-3: Zero-shot one image to 3d object.
\newblock In \emph{{IEEE/CVF} International Conference on Computer Vision, {ICCV} 2023, Paris, France, October 1-6, 2023}, pp.\  9264--9275. {IEEE}, 2023.
\newblock \doi{10.1109/ICCV51070.2023.00853}.
\newblock URL \url{https://doi.org/10.1109/ICCV51070.2023.00853}.

\bibitem[Mildenhall et~al.(2020)Mildenhall, Srinivasan, Tancik, Barron, Ramamoorthi, and Ng]{mildenhall2020nerf}
Mildenhall, B., Srinivasan, P.~P., Tancik, M., Barron, J.~T., Ramamoorthi, R., and Ng, R.
\newblock Nerf: Representing scenes as neural radiance fields for view synthesis.
\newblock In \emph{ECCV}, 2020.

\bibitem[Pag{\'e}s et~al.(2021)Pag{\'e}s, Amplianitis, Ondrej, Zerman, and Smolic]{pages2021volograms}
Pag{\'e}s, R., Amplianitis, K., Ondrej, J., Zerman, E., and Smolic, A.
\newblock Volograms \& v-sense volumetric video dataset.
\newblock \emph{ISO/IEC JTC1/SC29/WG07 MPEG2021/m56767}, 2021.

\bibitem[Radford et~al.(2021)Radford, Kim, Hallacy, Ramesh, Goh, Agarwal, Sastry, Askell, Mishkin, Clark, et~al.]{radford2021learning}
Radford, A., Kim, J.~W., Hallacy, C., Ramesh, A., Goh, G., Agarwal, S., Sastry, G., Askell, A., Mishkin, P., Clark, J., et~al.
\newblock Learning transferable visual models from natural language supervision.
\newblock In \emph{International conference on machine learning}, pp.\  8748--8763. PMLR, 2021.

\bibitem[Reimat et~al.(2021)Reimat, Alexiou, Jansen, Viola, Subramanyam, and Cesar]{reimat2021cwipc}
Reimat, I., Alexiou, E., Jansen, J., Viola, I., Subramanyam, S., and Cesar, P.
\newblock Cwipc-sxr: Point cloud dynamic human dataset for social xr.
\newblock In \emph{Proceedings of the 12th ACM Multimedia Systems Conference}, pp.\  300--306, 2021.

\bibitem[Ren et~al.(2023)Ren, Pan, Tang, Zhang, Cao, Zeng, and Liu]{rendreamgaussian}
Ren, J., Pan, L., Tang, J., Zhang, C., Cao, A., Zeng, G., and Liu, Z.
\newblock Dreamgaussian4d: Generative 4d gaussian splatting.
\newblock \emph{CoRR}, abs/2312.17142, 2023.
\newblock \doi{10.48550/ARXIV.2312.17142}.
\newblock URL \url{https://doi.org/10.48550/arXiv.2312.17142}.

\bibitem[Voleti et~al.(2024)Voleti, Yao, Boss, Letts, Pankratz, Tochilkin, Laforte, Rombach, and Jampani]{sv3d}
Voleti, V., Yao, C., Boss, M., Letts, A., Pankratz, D., Tochilkin, D., Laforte, C., Rombach, R., and Jampani, V.
\newblock {SV3D:} novel multi-view synthesis and 3d generation from a single image using latent video diffusion.
\newblock \emph{CoRR}, abs/2403.12008, 2024.
\newblock \doi{10.48550/ARXIV.2403.12008}.
\newblock URL \url{https://doi.org/10.48550/arXiv.2403.12008}.

\bibitem[Yoon et~al.(2021)Yoon, Yu, Park, and Park]{yoon2021humbi}
Yoon, J.~S., Yu, Z., Park, J., and Park, H.~S.
\newblock Humbi: A large multiview dataset of human body expressions and benchmark challenge.
\newblock \emph{IEEE Transactions on Pattern Analysis and Machine Intelligence}, 45\penalty0 (1):\penalty0 623--640, 2021.

\bibitem[Zhang et~al.(2022)Zhang, Zhang, Lin, Louie, and Huang]{zhangsim2real}
Zhang, T., Zhang, K., Lin, J., Louie, W.-Y.~G., and Huang, H.
\newblock Sim2real learning of obstacle avoidance for robotic manipulators in uncertain environments.
\newblock \emph{IEEE Robotics and Automation Letters}, 7\penalty0 (1):\penalty0 65--72, 2022.
\newblock \doi{10.1109/LRA.2021.3116700}.

\bibitem[Zhao et~al.(2023)Zhao, Yan, Xie, Hong, Li, and Lee]{zhao2023animate124}
Zhao, Y., Yan, Z., Xie, E., Hong, L., Li, Z., and Lee, G.~H.
\newblock Animate124: Animating one image to 4d dynamic scene.
\newblock \emph{arXiv preprint arXiv:2311.14603}, 2023.

\end{thebibliography}
\bibliographystyle{icml2024}

\newpage
\appendix
\onecolumn
\section{Implementation}
\label{appendix:implementation}

We open-source our implementation \href{https://github.com/rishab-partha/Vid3D}{here}, which provides training scripts for running Vid3D on an 8xA100 node.
\section{Related Work}
\label{sec:related_work}

Efficiently generating 3D scenes from 2D inputs has been a long-standing goal in computer vision~\citep{shapenet2015}. 
To automate and thus scale the building of static 3D scenes, researchers have recently leveraged advances in generative 2D video models to generate 3D scenes by querying a video model for multiple poses of said scene~\citep{chen2024v3d, han2024vfusion3d}.
However, these existing methods only generate static 3D data, meaning that there is no temporal component.

\paragraph{Domain-Specific Datasets}
Some existing highly domain-specific 3D video datasets (i.e. human bodies, human faces, etc.) have also been manually curated using a large number of sensors~\citep{chatzitofis2020human4d, reimat2021cwipc, yoon2021humbi, pages2021volograms}.
However, the use of these dynamic 3D scenes is limited due to their domain-specific nature.
Additionally, it is hard to scale their hardware-intensive video capture technology to produce more scenes.

\paragraph{Sim2Real}
Outside of 3D and 4D video datasets, other works focus on developing sim2real worlds, where robots learn real-world features in simulation~\citep{zhangsim2real,duautosim2real}. These works develop worlds where robots can interact in 3D space over time, but these 4D worlds require physics and temporal correlations to be designed by hand, hence not being able to generate dynamic scenes directly from a sample image.

\paragraph{Dynamic 3D Scenes}
Finally, a number of papers have focused on synthesizing dynamic 3D scenes either by using classifier guidance~\citep{bahmani4dfy, zhao2023animate124} or by training models that learn how to deform 3D objects~\citep{rendreamgaussian}. 
As stated before, these methods either suffer from high computational cost taking hours to render a single scene, or from rigid assumptions about the temporal structure of 3D scenes, limiting these methods' applicability for modeling arbitrary dynamic 3D scenes.
Our work instead leverages the strong generalization capabilities of 2D video models to flexibly generate 3D scenes without the need to train complex auxiliary models.

\section{Varying Scene Motion}
\label{appendix:ablate-motion}

\begin{table}[h!]
\centering
\caption{CLIP-I score values for Vid3D for different temporal seed motion scores. This result shows that there is a slight loss in quality for scenes with more motion.}
\vskip 0.15in
\label{tab:ablate-motion}
\begin{tabular}{lc}
\hline
\textbf{Motion Score} & \textbf{CLIP-I} \\ \hline
120 (Baseline) & 0.8946 \\
160 & 0.8893 \\
200 & 0.8897 \\
\hline
\end{tabular}
\vskip -0.1in
\end{table}

We ablate how the degree of motion in the scene's temporal seed (i.e., 2D video outline) affects the quality of the resulting 3D generations.
The model we use for temporal seeding, Stable Video Diffusion, generates videos conditioned on a "motion score", which corresponds to the time and spatial average of the optical flow maps of the video.
We thus vary the degree of motion in temporal seeds by varying the motion score.

We quantitatively measure the impact of varied motion in terms of CLIP-I score in~\Cref{tab:ablate-motion}.
While the initial increase of conditioned motion score from 120 to 160 degrades the CLIP-I score of the renderings by 0.0053, there is almost no difference between the CLIP-I metrics for motion scores of 160 and 200.
This result suggests that while there is some initial loss in quality due to increased motion, past a given motion score the quality of Vid3D is relatively robust to added motion.

\begin{figure}[htbp]
    \centering
    \includegraphics[width=0.8\linewidth]{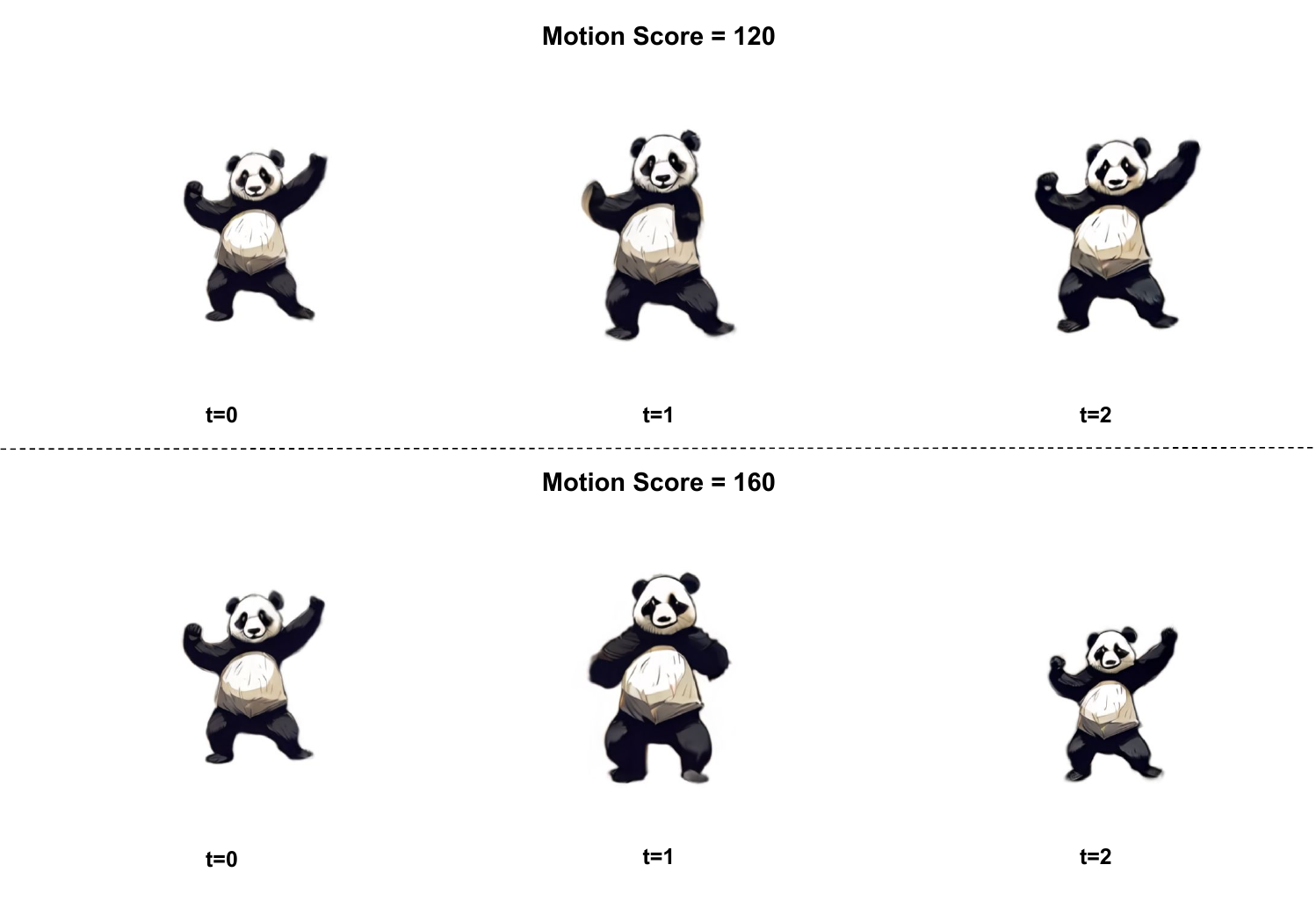}
    \caption{Rendering of singular frame from 3D videos generated by the same reference image but with different amount of motion synthesized. As desired, the higher motion score video has higher variability, along with similar rendering quality to the lower motion score, demonstrating robustness to motion.}
    \label{fig:motion-ablation}
\end{figure}

We provide a qualitative example of varying the motion score in~\cref{fig:motion-ablation}.
For each motion score, we render a 2D video from the 3D video using a fixed camera view.
We observe greater variability in poses between frames for the higher motion score.
Furthermore, the added motion is achieved without a noticeable drop in rendering quality.
\section{Qualitative Analysis}
\label{appendix:qualitative}

\begin{figure}[htbp]
    \centering
    \includegraphics[width=\linewidth]{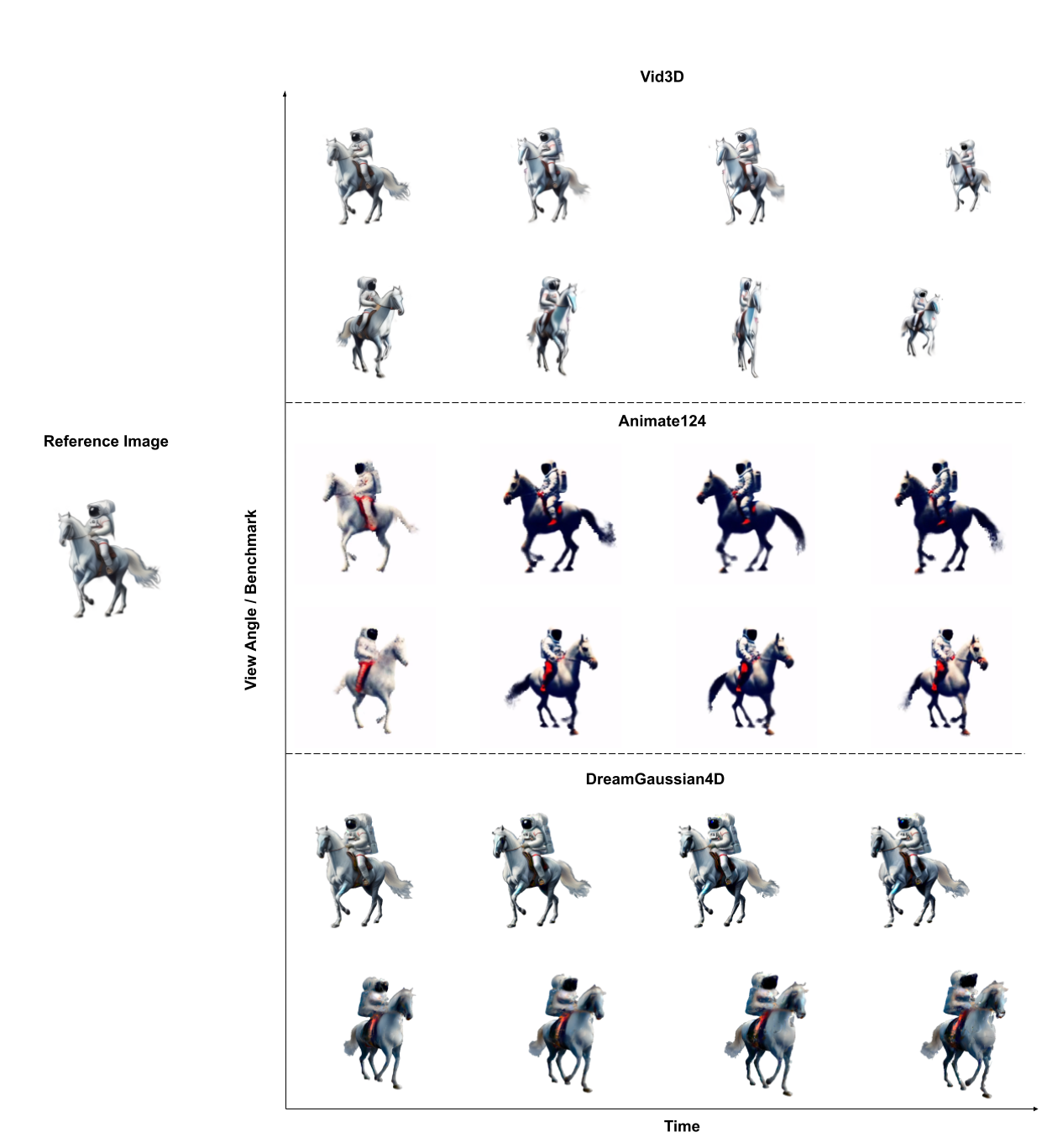}
    \caption{A qualitative comparison of Animate124, DreamGaussian4D, and Vid3D for a seed image of an astronaut riding a horse. Here, Vid3D both creates accurate representations from multiple angles, but also does not recolor the horse like Animate124 or have worse renders from a non-reference view like DreamGaussian4D.}
    \label{fig:comparison-astronaut}

\end{figure}

\begin{figure}[htbp]
    \centering
    \includegraphics[width=\linewidth]{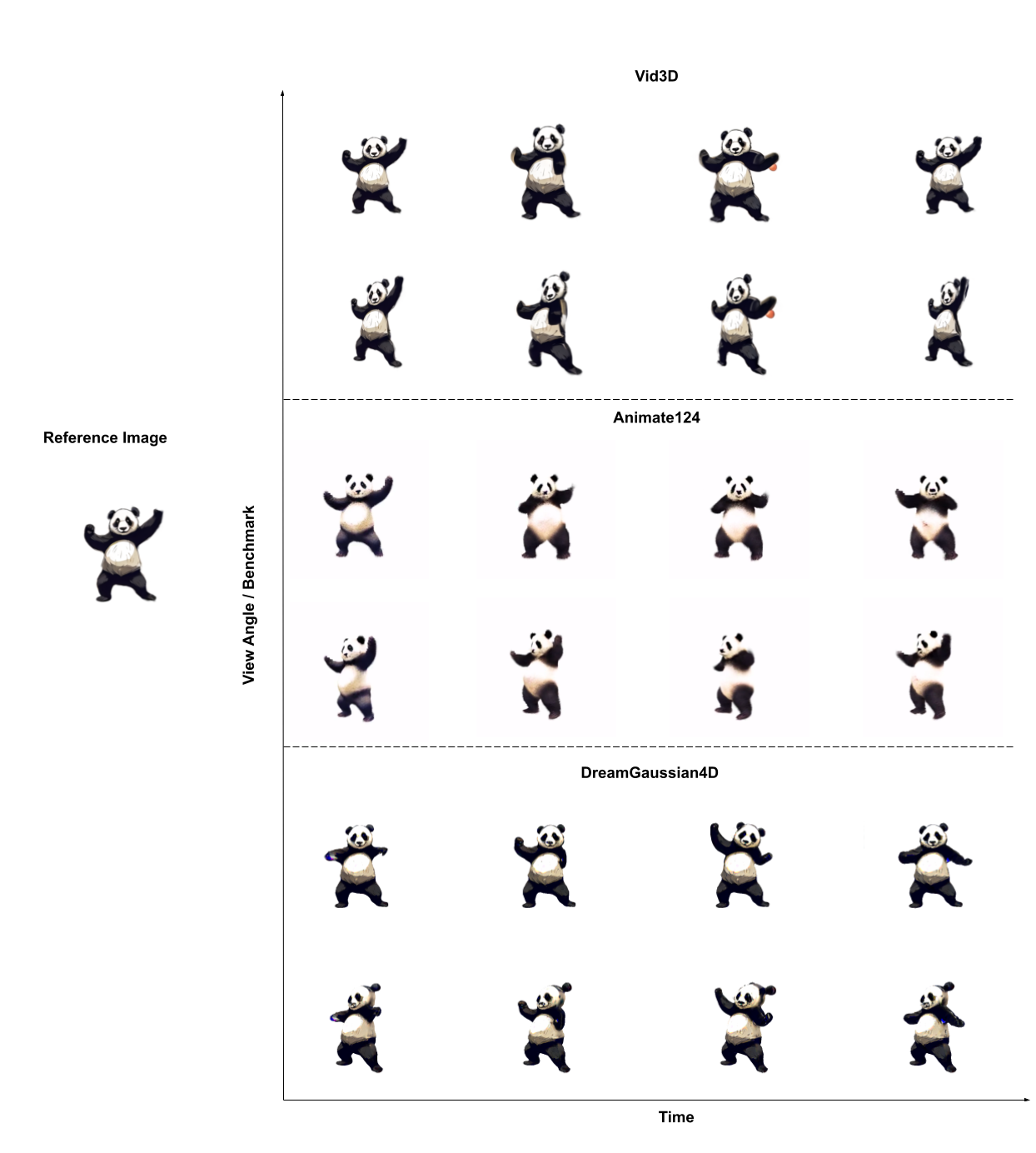}
    \caption{A qualitative comparison of Animate124, DreamGaussian4D, and Vid3D for a seed image of a dancing panda. Here, although Vid3D includes the hallucination of a orange ball (discussed in Figure 10), in comparison to DreamGaussian4D, Vid3D does not hallucinate only one ear to the panda.}
    \label{fig:comparison-panda}

\end{figure}

\begin{figure}[htbp]
    \centering
    \includegraphics[width=\linewidth]{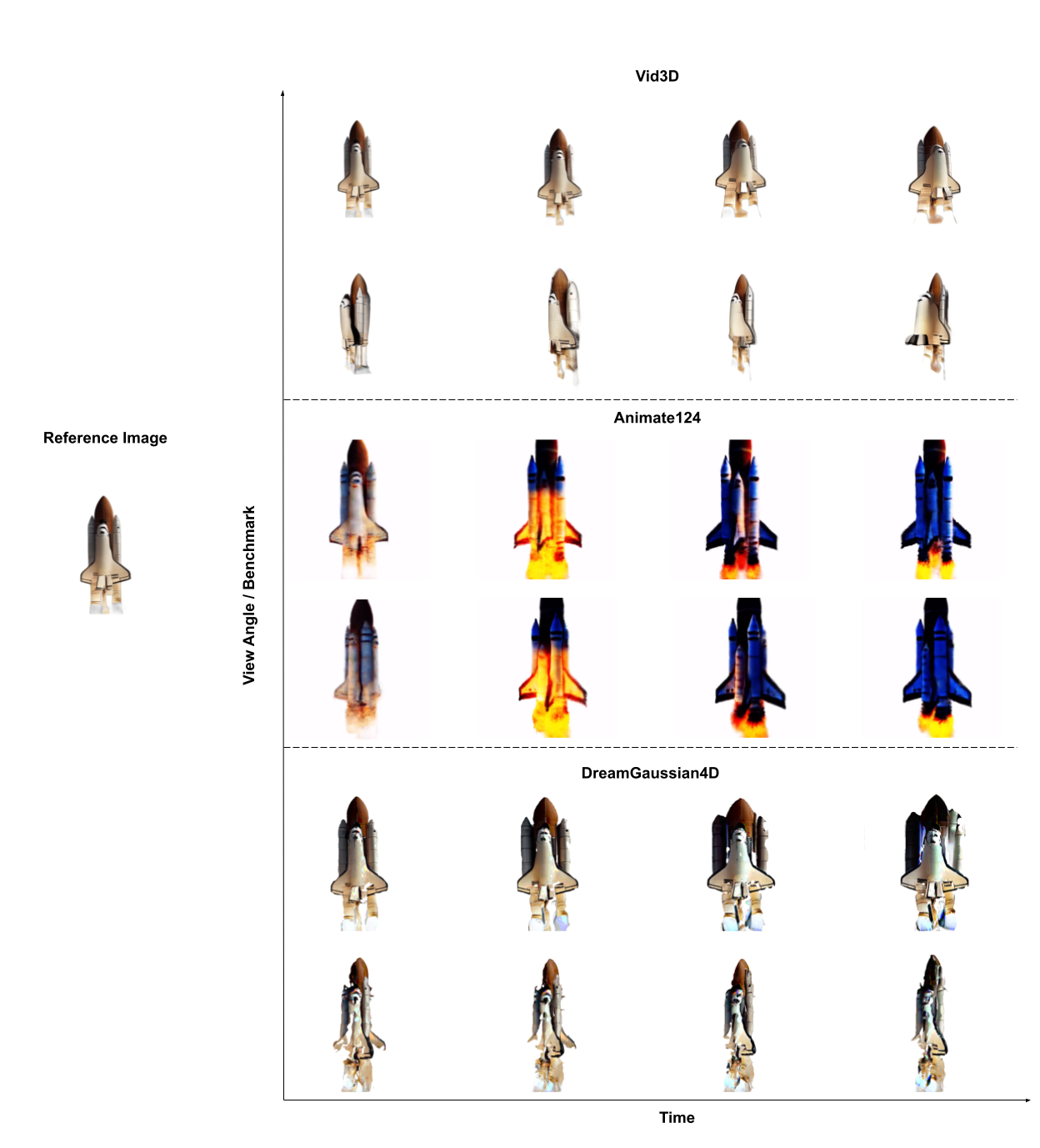}
    \caption{A qualitative comparison of Animate124, DreamGaussian4D, and Vid3D for a seed image of a space shuttle. Here, the quality of the exhaust is higher in Vid3D, with the exhaust recoloring the rocket in Animate124 and the space shuttle being deformed in the side angle in DreamGaussian4D.}
    \label{fig:comparison-rocket}

\end{figure}

\begin{figure}[htbp]
    \centering
    \includegraphics[width=\linewidth]{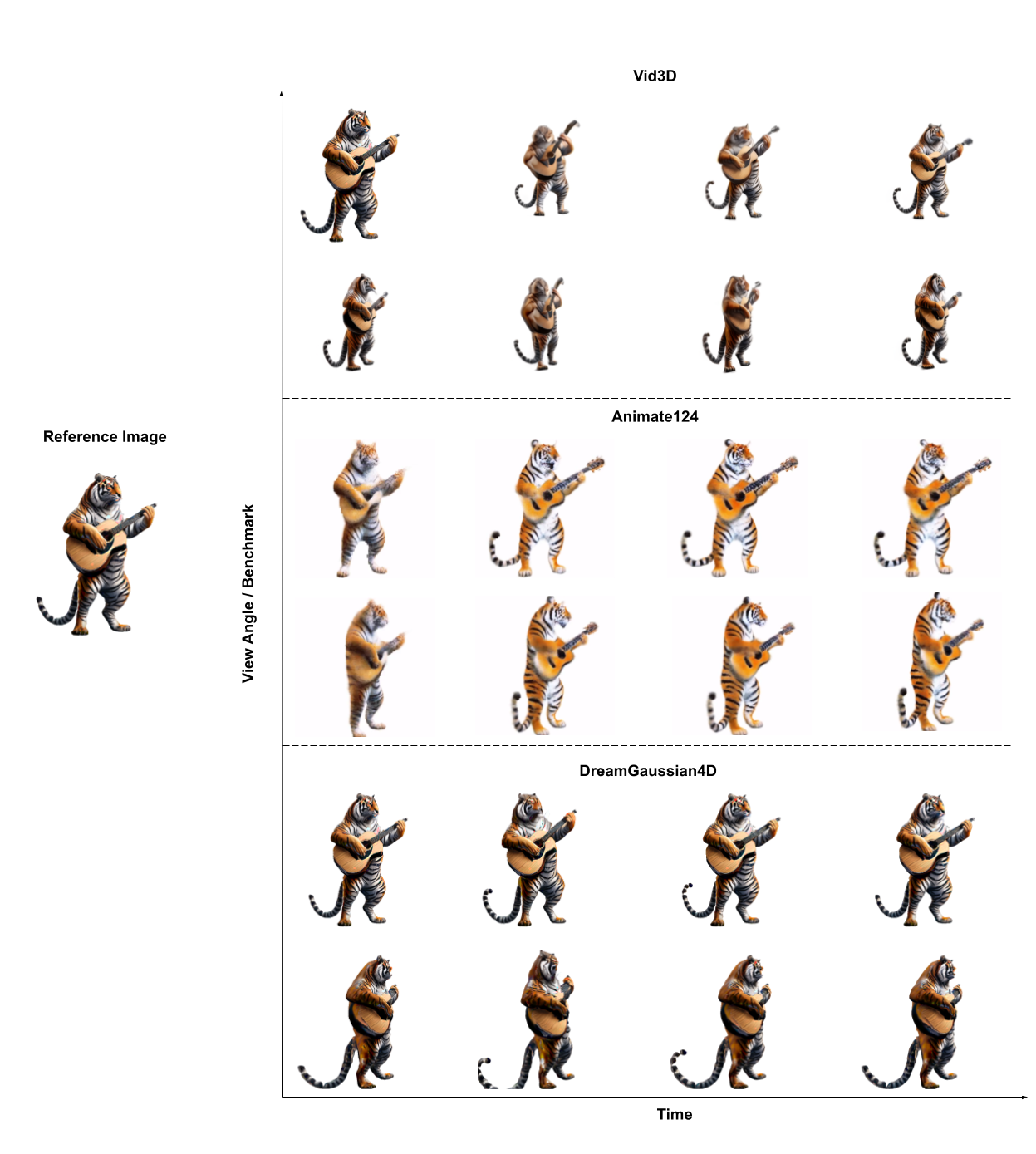}
    
    \caption{A qualitative comparison of Animate124, DreamGaussian4D, and Vid3D a seed image of a tiger playing the guitar. Here, we notice that our method has higher coherence to the source image than Animate124, keeping the same style and not distorting the shape of the tiger like DreamGaussian4D.}
    \label{fig:comparison-tiger}
\end{figure}

We compare the performance qualitatively of Vid3D to DreamGaussian4D and Animate124 on four examples~\citep{rendreamgaussian, zhao2023animate124} in ~\Cref{fig:comparison-astronaut}, ~\Cref{fig:comparison-panda}, ~\Cref{fig:comparison-rocket}, and ~\Cref{fig:comparison-tiger}. We note that only four examples are chosen because Animate124 requires 10 hours of textual inversion and 7 hours of classifier guidance in order to produce a singular video, in comparison to the ten minutes from our method.

We observe that Vid3D has high quality generated 3D videos on multiple axes. First, the video hallucinates less than Animate124, which frequently changes the color and appearance of objects, such as in the space shuttle and the astronaut, along with higher resolution. Second, DreamGaussian4D has high quality reference-view renders but struggles on alternate views, such as those for the space shuttle. On the other hand, Vid3D achieves high performance over all axes. 

\begin{figure}[htbp]
    \centering
    \includegraphics[width=\linewidth]{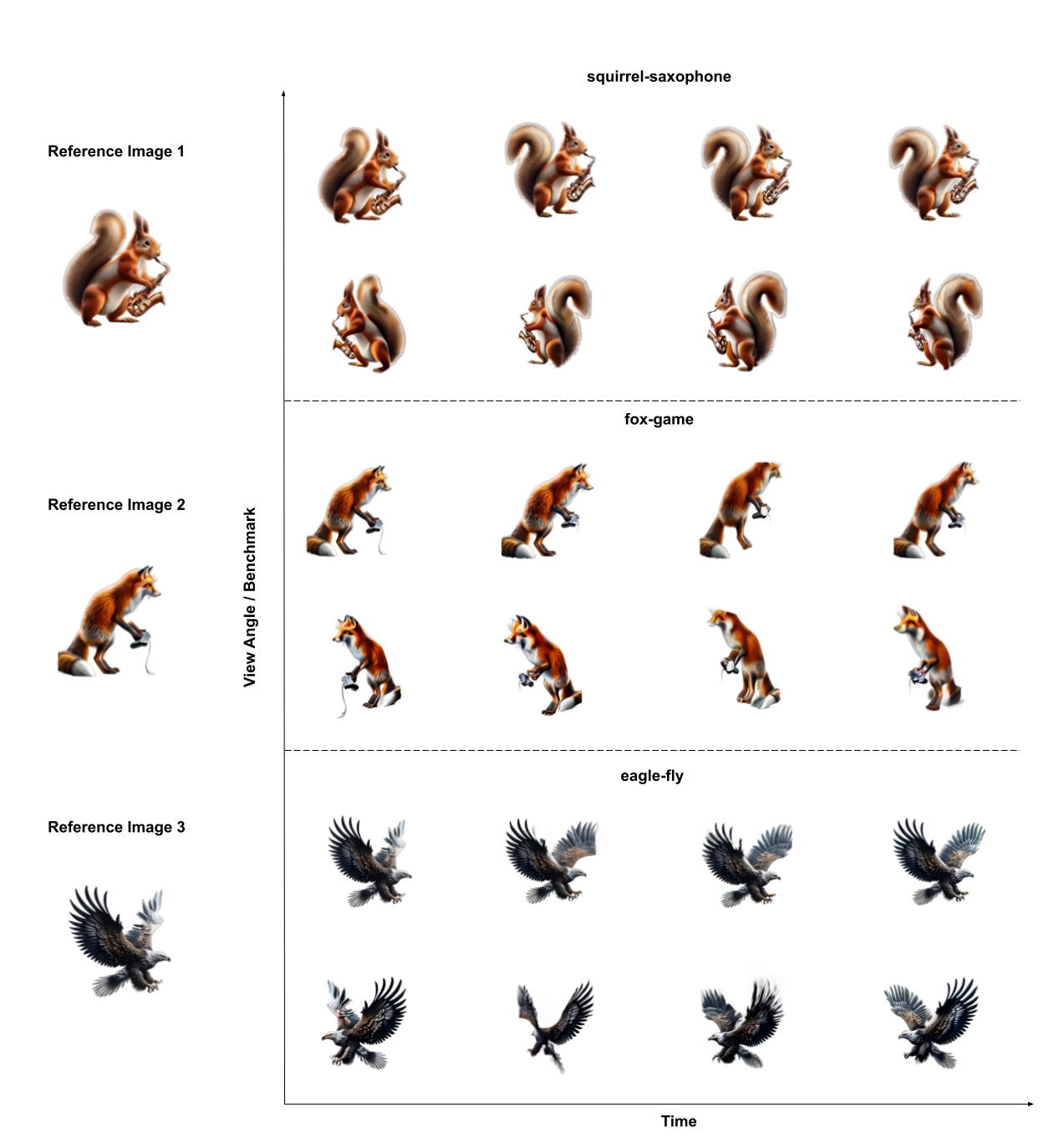}
    \caption{Further examples of high quality generations from Vid3D, demonstrating generalization over multiple timesteps and views.}
    \label{fig:examples}

\end{figure}

In ~\Cref{fig:examples}, we present further examples of generations from Vid3D, which demonstrate strong coherence both temporally and over multiple views. This demonstrates that Vid3D can learn strong 3D temporal priors even though it solely relies on 2D temporal priors and 2D video diffusion models.

\subsection{Failure Modes}

\begin{figure}[htbp]
    \centering
    \includegraphics[width=\linewidth]{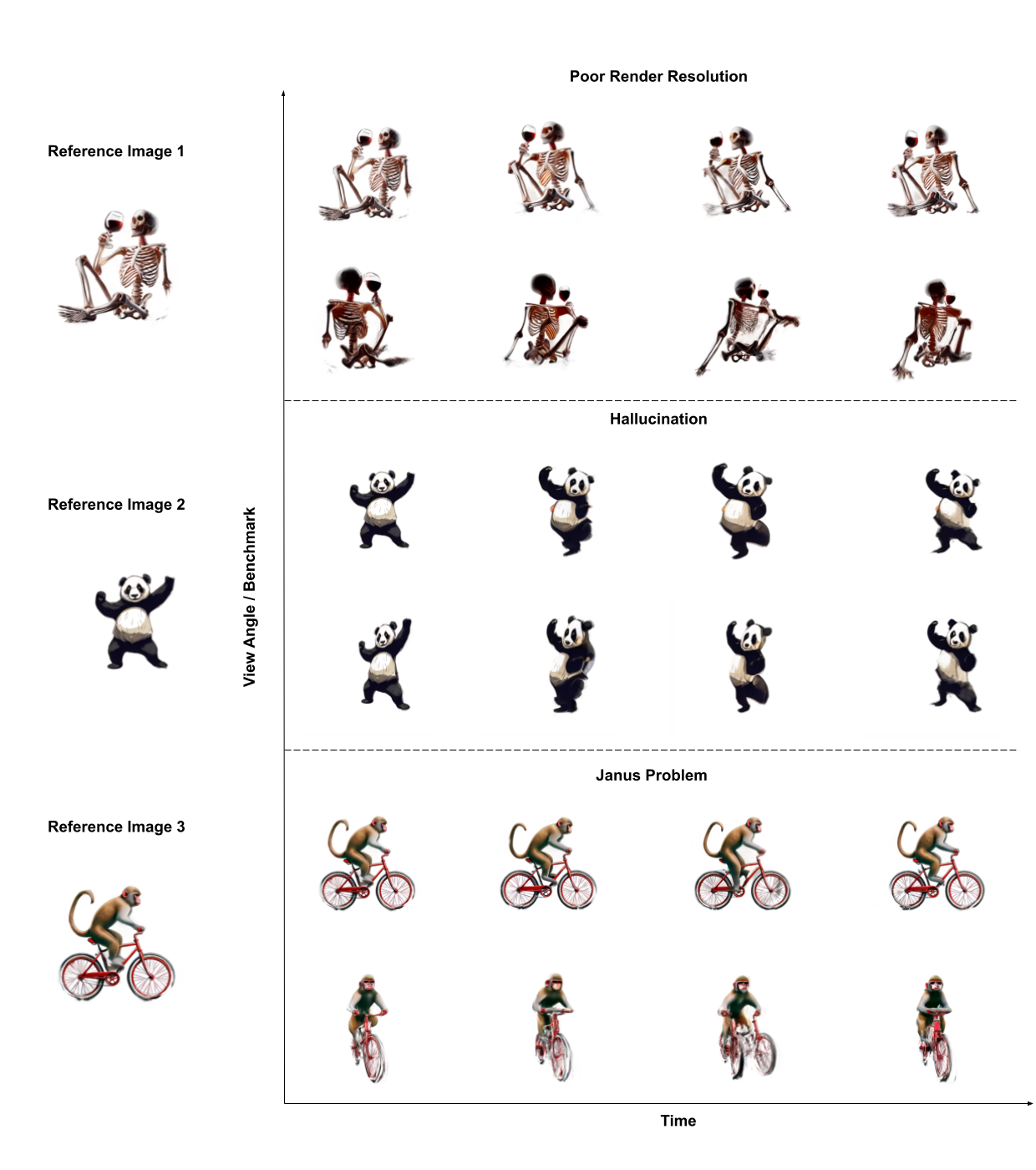}
    \caption{A depiction of various failure modes for Vid3D. The first is poor render quality, where the skeleton has its facial features blurred due to the failure of the 3D modeling to appropriately represent it. The second is hallucination, where a orange ball appears near the panda. The final is the Janus problem, where the same object appears twice, which is a known problem in 3D generation and once again appears~\citep{janus}.}
    \label{fig:failure-modes}
\end{figure}

We address various failure modes of Vid3D in ~\Cref{fig:failure-modes}. The first is a lack of quality in the generation of multi-views, which leads to issues like the face of the skeleton not being properly rendered. We plan to hopefully address this by potentially training our own multi-view generation model. The next issue is hallucination, where new objects, like the orange ball are added to existing scenes. This issue arises because of the flexibility of our algorithm--we do not simply deform the scene. However, this means that occassionally, small aberrations like the orange ball may appear. Finally, we encounter the Janus problem common in 3D generation~\citep{janus}. For example, in the third frame of the bike, the bike appears twice, which is an issue that can only be addressed through better training of multi-view models.

These issues are all common issues faced by 3D generation models, and do not take away from the value of this experiment. We thus still maintain that 3D temporal knowledge may not be necessary for 3D video generation.


\end{document}